\documentclass[conference]{IEEEtran}
\IEEEoverridecommandlockouts

\usepackage{cite}
\usepackage{amsmath,amssymb,amsfonts}
\usepackage{algorithmic}
\usepackage{graphicx}
\usepackage{textcomp}
\usepackage{xcolor}
\def\BibTeX{{\rm B\kern-.05em{\sc i\kern-.025em b}\kern-.08em
    T\kern-.1667em\lower.7ex\hbox{E}\kern-.125emX}}
\begin{document}
\bstctlcite{IEEEexample:BSTcontrol}

\title{Sim-to-Real RL for ASVs using SysID \\
}


\author{
\IEEEauthorblockN{Cody Sheltraw\textsuperscript{1}, Tsimafei Lazouski\textsuperscript{1}, Maani Ghaffari\textsuperscript{1}, and Alan Papalia\textsuperscript{1}}
\IEEEauthorblockA{\thanks{\textsuperscript{1} University of Michigan, Ann Arbor, MI 48109, USA. Corresponding author e-mail: \texttt{cshel@umich.edu}}
}}

\maketitle

\begin{abstract}
Autonomous Surface Vehicles (ASVs) operating in dynamic marine environments require robust control policies for tasks such as path following and station keeping, making reinforcement learning (RL) a promising alternative to classical controllers.
However, existing ASV simulators rarely support parallel environments for RL training.
Such existing simulators require accurate hydrodynamic modeling from computational fluid dynamics solvers or towing tank tests for setting hydrodynamic parameters to address the sim-to-real gap.
To address these challenges, we present an ASV simulator and accompanying pipeline that enables training policies starting from unknown vehicle dynamics.
Our framework uses only a CAD model and brief set of open-water field trajectories for approximating and refining both hydrodynamic and thruster parameters.
Real-world deployments on a BlueBoat ASV demonstrate successful zero-shot sim-to-real transfer in path following and station-keeping tasks without prior hydrodynamic and propeller information.
\end{abstract}


\section{Introduction}

Autonomous surface vehicles (ASVs) operating in marine environments require robust control despite non-linear hydrodynamic interactions, uncertain propulsion characteristics, and environmental disturbances, such as wind, waves, and currents~\cite{xu2022trajectorytracking,santos2023robustcontrol}. 
Since these hydrodynamic forces depend directly on the vehicle's operating state and surrounding sea conditions, 
accurate modeling and controller tuning are challenging for tasks such as path following and station keeping.
Reinforcement learning (RL) offers an attractive alternative for such tasks by learning control policies directly from interactions, without explicit modeling of each relevant dynamic interaction, necessitating the availability of RL-capable simulators. In other robotic domains, RL policies have demonstrated the ability to control complex non-linear systems in challenging operating conditions~\cite{agileleggedhumanoid2019, liu2025discretetimelearning}.
However, transferring policies trained in simulation to real-world deployment remains challenging for ASVs due to the sim-to-real gap between simulated and real marine dynamics~\cite{batista2024ASVDRL,batista2025evaluatingDRL}.

To reduce this sim-to-real gap, an adequate representation of the vessel dynamics is needed.
Such high-fidelity models traditionally rely on towing tank experiments or computational fluid dynamics (CFD) solvers for accurate hydrodynamic system identification (SysID), creating a high barrier to entry. Furthermore, using real-world SysID parameters in a simulator does not guarantee improved sim-to-real performance~\cite{chebotar2019closingsimtorealloopadapting}. The mismatch between real-world and simplified simulator dynamics means parameters identified from the real vessel may not align with the simulator parameters that best reproduce its observed behavior.
To compensate for this, other works in the marine domain rely on heavy domain randomization (DR), which can lead to conservative policies~\cite{xu2022groupdistributionallyrobustRL} and decreased real-world performance~\cite{nachum2019multiagentmanipulation}.

To address these limitations, this work presents a simulator and an accompanying real-to-sim-to-real pipeline that facilitates the deployment of simulation-trained policies on a physical ASV using little real-world data.
Our approach reduces the sim-to-real gap by grounding the simulated dynamics in observed real-world ASV behavior using the covariance matrix adaptation evolution strategy (CMA-ES) over a set of trajectories.
Starting with unknown vessel dynamics, our method reduced station-keeping position error by $63.9\%$ (1.074~m to 0.388~m RMSE) and heading error by $40.1\%$ (17.30$^\circ$ to 10.37$^\circ$ MAE) over an analytical SysID approach under varying currents on a physical BlueBoat~\cite{blueboat}.

\begin{figure}
    \centering
    \includegraphics[width=1.0\linewidth]{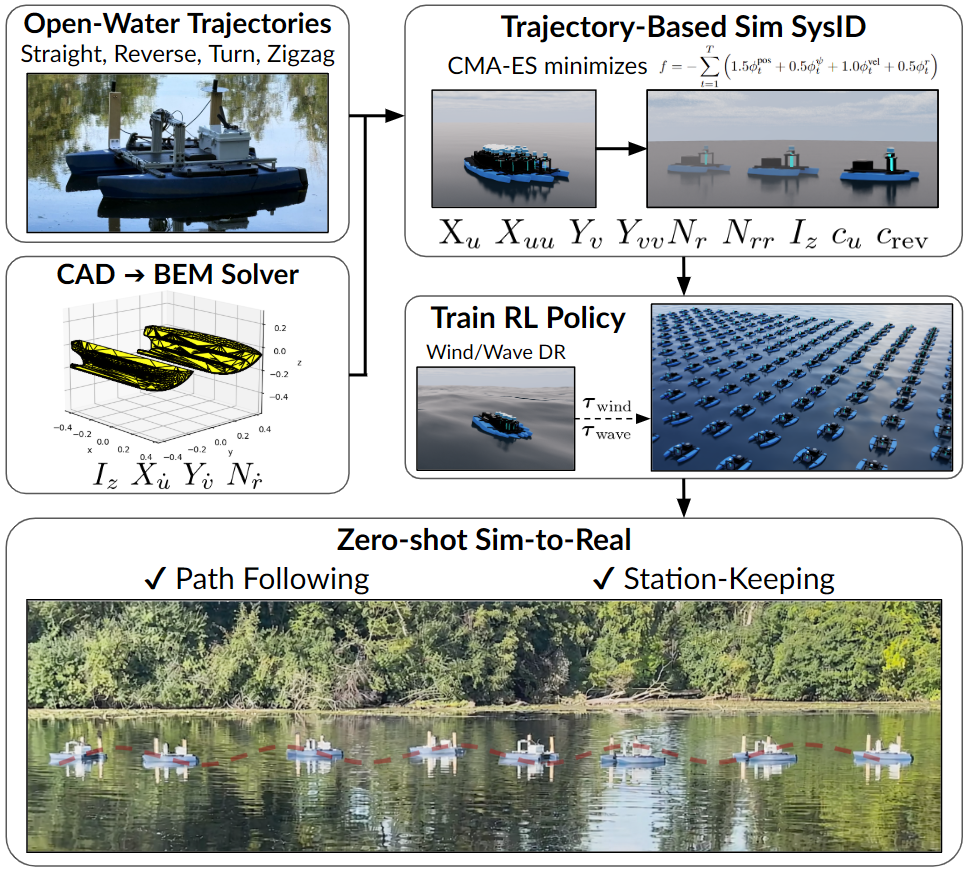}
    \caption{Overview of our sim-to-real pipeline. Initial hydrodynamic parameters are derived from a hull CAD model via a boundary element method (BEM) solver alongside a set of open-water trajectories from the physical autonomous surface vehicle (ASV). System identification (SysID) is performed in the simulator using the covariance matrix adaptation evolution strategy (CMA-ES) to obtain refined hydrodynamic ($X_u, X_{uu}, Y_v, Y_{vv}, N_r, N_{rr}, I_z$) and thruster ($c_u, c_{\text{rev}}$) parameters, grounding the simulator to the real ASV. Reinforcement learning (RL) policies are trained in the grounded simulator, enabling zero-shot transfer for path following and station-keeping tasks.}
    \label{fig:sys_arch}
\end{figure}

The main contributions of this work are as follows:
\begin{enumerate}
    \item \textbf{ASV Simulator}: An RL-focused ASV simulator based on NVIDIA Isaac Sim with hydrodynamics, wind-wave disturbance, and propeller-thrust modeling.
    \item \textbf{Accessible real-to-sim pipeline}: A CAD model and a single set of open-water trajectories are used to ground simulated vessel behavior, eliminating the costly need for towing tanks or CFD solvers.
    \item \textbf{Field sim-to-real evaluation}: Station-keeping and path following policies trained using our simulator and methodology are deployed and evaluated on a real ASV in a river environment.
\end{enumerate}




\section{Related Work}

\subsection{Marine Simulation}
Several marine simulators with RL capabilities are available today, including one that was built specifically for the ASV application, and whose policies have been deployed in the field~\cite{batista2024ASVDRL,batista2025evaluatingDRL,wang2023DRLasv}. However, it is strictly limited to planar dynamics, not accounting for roll or pitching motions, and constant force and torque are used to simulate water currents. Other notable RL marine simulators focus on the underwater domain~\cite{potokar2022holoocean, stonefish, chu2025marinegym}, and adaptations of these simulators for surface vehicles are still in development.
Our simulator, exclusively built for surface vehicles, is developed using real-world field data, separating it from existing purely analytical solutions. To our knowledge, this is also the first ASV simulator that supports parallel environments for RL with geometric wave displacement updating the submerged volume of an ASV.

\subsection{Sim-to-Real}

Domain randomization (DR) and grounded simulation learning (GSL)~\cite{farchy2013GSL} are two proven methods used to reduce the sim-to-real gap.

\subsubsection{Domain Randomization}
Domain randomization (DR) adds
variations to the space of simulated environments to prevent trained policies
from overfitting \cite{tobin2017domainrandomization}. DR has been used extensively in the field of robotics for the
zero-shot sim-to-real transfer of
drones~\cite{loquercio2020DeepDroneRacing,andres2025quadcopterwind},
legged robotics~\cite{liao2025beyondmimic}, and
ASVs~\cite{richard2021trainheron,batista2024ASVDRL}. While heavy DR
improves generalization, even for unmodeled
disturbances~\cite{andres2025quadcopterwind}, the resulting policies tend to be
conservative~\cite{xu2022groupdistributionallyrobustRL}. 
%
%
Therefore, heavy DR should primarily be used for unmodeled dynamics and unknown disturbances, such as wind and water in the marine domain, instead of parameters with measured real-world values.


\subsubsection{Grounded Simulation Learning}
GSL is an iterative process of using real-world data to ground a simulator, deploying the policy in the real world, and then refining simulation parameters based on expected dynamics versus real telemetry. While multiple iterations provide slight improvements over time for legged robots~\cite{farchy2013GSL}, they assume zero environmental disturbance over the same task, which makes multiple iterations unrealistic for the marine domain. Grounded Action Transformation (GAT)~\cite{hanna2017GAT} is another framework extending GSL, but instead treats the simulator as a black box. A task-specific transformation mapping is learned to match the system's actions in the source domain to actions in a target domain. Since we can modify the hydrodynamics of our simulator, we can perform task-agnostic grounding without GAT.



\section{Methodology}
Our methodology is split into three phases: (1) formulation of computationally efficient hydrodynamics, environmental disturbances, and propeller thrust modeling in simulation; (2) simulator parameter fitting using real-world SysID and trajectory data; and (3) Markov Decision Process (MDP) and DR formulations to support the zero-shot sim-to-real transfer of a trained policy to a physical ASV.

\subsection{Simulation and Environment Design}
This project uses NVIDIA Isaac Lab 2.3.2~\cite{nvidia2025isaaclab} built on Isaac Sim 5.1.0~\cite{NVIDIA_Isaac_Sim}, using Omniverse PhysX for physics.

\subsubsection{Hydrodynamics}
The ASV is natively modeled as a 6-degree-of-freedom (6-DOF) system in PhysX, while SysID and maneuvering are modeled in 3-DOF based on Fossen's relative motion model~\cite{fossen2021}:
\begin{equation}
    M\dot{\boldsymbol{\nu}}_\text{r} + C(\boldsymbol{\nu}_\text{r})\boldsymbol{\nu}_\text{r} + D(\boldsymbol{\nu}_\text{r})\boldsymbol{\nu}_\text{r} = \boldsymbol{\tau} + \boldsymbol{\tau}_{\text{wave}} + \boldsymbol{\tau}_{\text{wind}}
    \label{eq:fossen_3dof}
\end{equation}
where $\boldsymbol{\nu}_\text{r} = \boldsymbol{\nu} - \boldsymbol{\nu}_\text{c} = [u_\text{r}, v_\text{r}, r]^\top$ is the relative body velocity vector between the vehicle velocity $\boldsymbol{\nu}$ and a water current velocity $\boldsymbol{\nu}_\text{c}$. $M = M_{\text{RB}} + M_{\text{A}}$ accounts for rigid-body mass and added mass inertia. The matrices $C(\boldsymbol{\nu}_\text{r})$ and $D(\boldsymbol{\nu}_\text{r})$ account for the Coriolis-centripetal and hydrodynamic damping effects, respectively. $\boldsymbol{\tau}$ represents the propulsion wrench, while $\boldsymbol{\tau}_{\text{wave}}$ and $\boldsymbol{\tau}_{\text{wind}}$ represent wave and wind-induced forces, respectively. Two simplifications are made: (1) port-starboard geometric symmetry is assumed, eliminating surge cross-coupling coefficients, and (2) off-diagonal damping terms in $D(\boldsymbol{\nu}_\text{r})$ are neglected in the low-speed operating regime.

The wave implementation is intended for semi-realistic yet computationally efficient reinforcement learning rather than high-fidelity CFD. We incorporate a hydrodynamic wave model using NVIDIA Warp~\cite{macklin2022warp}, following an internal Isaac Sim 5.1.0 example ocean~\cite{NVIDIA_Isaac_Sim} based on empirical directional wave spectra defined in~\cite{horvath2015wavespectra}. Simulated waves are derived using the Texel-Marsen-Arsloe (TMA) spectrum~\cite{TMA_spectrum}, which generalizes the Joint North Sea Wave Project (JONSWAP) spectrum~\cite{DirectionalWaveSpectraJONSWAP1980} to finite water depths.

The sea surface elevation spectrum $S_{\text{TMA}}(\omega)$ is defined as:
\begin{equation}
    S_{\text{TMA}}(\omega) = S_{\text{JONSWAP}}(\omega) \Phi(\omega,h)
    \label{eq:tma_spectrum}
\end{equation}
where $\omega$ is the wave frequency, $S_{\text{JONSWAP}}(\omega)$ is the fetch-limited JONSWAP spectrum~\cite{DirectionalWaveSpectraJONSWAP1980} and $\Phi(\omega,h)$ is the Thompson--Vincent~\cite{thompson1983prediction} depth-attenuation factor, evaluated from the dimensionless depth $\omega_h = \omega\sqrt{h/g}$. The JONSWAP spectrum follows a Pierson-Moskowitz form with peak-enhancement factor $\gamma=3.3$, Phillips parameter $\alpha$, and peak frequency $\omega_p$ set by prescribed wind speed and fetch.

The buoyant forces are governed by hydrostatic principles, with the buoyant force proportional to the submerged hull volume $V_{\text{sub}}$. The simulator evaluates $V_{\text{sub}}$ in real-time using geometry-based water surface interactions. During each physics step, the integrated draft $D_j$ for each hull $j \in \{1, \dots, N_{\text{hulls}}\}$ is computed across its $M_v$ surface vertices:
\begin{equation}
    D_j = \sum_{i=1}^{M_v} \max\bigl(0, \min(d_{\max,i}, h_{w,i} - z_i)\bigr)
\label{eq:buoyancy}
\end{equation}
where $d_{\max,i}$ is the pre-computed maximum draft, $h_{w,i}$ is the local wave height, and $z_i$ is the vertical position of vertex $i$.

The summed draft is converted to a submerged volume $V_{\text{sub},j}$ through a piecewise-linear displacement coefficient $\kappa_j$:
\begin{equation}
    \kappa_j =
    \begin{cases}
        \dfrac{V_{\text{rest},j}}{D_{\text{rest},j}} & \text{if } D_j \le D_{\text{rest},j} \\[2.2ex]
        \dfrac{V_{\text{full},j} - V_{\text{rest},j}}{D_{\text{full},j} - D_{\text{rest},j}} & \text{if } D_j > D_{\text{rest},j}
    \end{cases}
    \label{eq:kappa}
\end{equation}
\begin{equation}
    V_{\text{sub},j} = V_{\text{rest},j} + \kappa_j\,(D_j - D_{\text{rest},j}), \qquad V_{\text{sub}} = \sum_j V_{\text{sub},j}
    \label{eq:v_sub}
\end{equation}
where $D_{\text{rest},j}$ and $D_{\text{full},j}$ are the integrated drafts at the resting waterline and full submergence, $V_{\text{rest},j}$ and $V_{\text{full},j}$ are the corresponding CAD-measured volumes, and $V_{\text{sub},j}$ is clamped to $[0, V_{\text{full},j}]$. The parameter $\kappa_j$ serves as an effective cross-sectional area converting draft depth to volume. The resulting buoyant force $F_\text{b} = \rho g\, V_{\text{sub}}$, with water density $\rho$ and gravity $g$, acts vertically at the centroid of each submerged hull section $j$, yielding restoring moments in heave, roll, and pitch. Body-frame velocities and wrenches computed from the hydrodynamic and hydrostatic forces are applied to the vehicle every simulation physics step.



\subsubsection{Wind}
Ambient wind is modeled as physical forces acting on the ASV via a generalized aerodynamic wrench model adapted from Fossen. Using the relative wind velocity components ($u_{\text{rw}}, v_{\text{rw}}$), the apparent wind speed and relative wind angle of attack are computed as $V_{\text{rw}} = \sqrt{u_{\text{rw}}^2 + v_{\text{rw}}^2}$ and $\gamma_{\text{rw}} = -\operatorname{atan2}(v_{\text{rw}}, u_{\text{rw}})$. The resulting wind wrench $\boldsymbol{\tau}_{\text{wind}}$ is then calculated as:
\begin{equation}
\boldsymbol{\tau}_{\text{wind}} = 
\begin{bmatrix}
-\frac{1}{2} \rho_a V_{\text{rw}}^2 C_X\cos(\gamma_{\text{rw}}) A_{\text{Fw}} \\
\frac{1}{2} \rho_a V_{\text{rw}}^2 C_Y\sin(\gamma_{\text{rw}}) A_{\text{Lw}} \\
0 \\
0 \\
0 \\
\frac{1}{2} \rho_a V_{\text{rw}}^2 C_N\sin(2\gamma_{\text{rw}}) L_{\text{OA}} A_{\text{Lw}}
\end{bmatrix}
\label{eq:tau_wind}
\end{equation}
where $\rho_a$ is air density, $A_{\text{Fw}}$ and $A_{\text{Lw}}$ are the frontal and lateral projected areas, $L_{\text{OA}}$ is the vessel overall length, and $C_X$, $C_Y$, and $C_N$ are the surge, sway, and yaw force and moment coefficients, respectively. For our BlueBoat with its modified sensor suite, these parameters are fixed at nominal values: $\rho_a = 1.225\text{ kg/m}^3$, $L_{\text{OA}} = 1.05\text{ m}$, $A_{\text{Fw}} = 0.115\text{ m}^2$, $A_{\text{Lw}} = 0.163\text{ m}^2$, $C_X = 0.90$, $C_Y = 0.20$, and $C_N = 0.30$.

\subsubsection{Propeller Thrust}
\label{sec:propeller_thrust}
Open-water propeller thrust $T$ is determined by the standard marine propeller thrust equation $T=k_T(J)\rho n^2 D^4$, where $\rho$ is the water density (kg/m$^3$), $n$ is the propeller rotational speed (rev/s), $D$ is the propeller diameter (m), and $k_T(J)$ is the thrust coefficient as a function of the advance ratio $J=\tfrac{u_a}{nD}$, with $u_a$ denoting the advance inflow velocity. For real-world system identification purposes, we derive coefficients $c_u$ and $c_{\text{rev}}$ from this standard model to capture advance loss and reverse efficiency, respectively. For simulation purposes, we introduce motor spool-up lag $\tau_{\text{lag}}$ (nominally $0.15$\,s) alongside a starboard motor imbalance factor $c_{\text{stbd}}$ (nominally $1.0$) aimed at capturing propeller performance asymmetries between the port and starboard thrusters.

Linearizing $k_T(J) \approx K_0 - K_1 J=K_0-K_1(\tfrac{u_a}{nD})$ around nominal advance ratios separates the static bollard pull $T_0$ from the advance inflow loss $\Delta T_u$:
\begin{subequations}
\begin{align}
    T_0 &= K_0 \rho D^4 n^2
    \label{eq:static_force} \\
    \Delta T_u &= K_1 \rho D^3 n u_a
    \label{eq:advance_loss}
\end{align}
\end{subequations}
Eliminating $n$ between~\eqref{eq:static_force} and~\eqref{eq:advance_loss} and evaluating the fractional loss $\Delta T_u/T_0$ at a nominal operating thrust $T_0 = T_{\text{n}}$ gives the dimensionless linearized advance-loss factor $c_u$:
\begin{equation}
    c_u = \frac{K_1 \sqrt{\rho} D}{\sqrt{K_0 T_{\text{n}}}}
    \label{eq:c_u}
\end{equation}
where $K_0$ is the zero-speed thrust coefficient, $K_1$ is the slope of $k_T(J)$, and $T_{\text{n}}$ is the nominal operating thrust. The net active thrust is thus modeled as $T \approx T_0(1 - c_u u_a)$.

Motor spool-up lag is modeled as a first-order low-pass filter applied to the commanded thrust $T_{\text{cmd}}$ with time constant $\tau_{\text{lag}}$:
\begin{equation}
    T_{\text{act}}[k] = T_{\text{act}}[k-1] + \alpha \bigl(T_{\text{cmd}}[k] - T_{\text{act}}[k-1]\bigr)
\label{eq:c_lag}
\end{equation}
where $T_{\text{act}}[k]$ is the actual active thrust, $T_{\text{cmd}}[k]$ is the target thrust command, $\alpha=\frac{\Delta t}{\tau_{\text{lag}}}$, and $\Delta t$ is the simulator's physics timestep. Note that $\Delta t\ll \tau_{\text{lag}}$, and $\tau_{\text{lag}}$ is utilized for dynamic simulation only. During steady-state and coasting SysID, sequential thrust commands are assumed to be equivalent ($T_{\text{cmd}}[k] = T_{\text{cmd}}[k-1] = T_{\text{act}}[k-1]$), eliminating motor lag.

Physical thrust asymmetries between port/starboard thrusters and forward/reverse operation are defined via dimensionless scaling ratios $c_{\text{stbd}}$ and $c_{\text{rev}}$:

\vspace{5pt}
\begin{subequations}
\noindent
\begin{minipage}{0.48\columnwidth}
\begin{equation}
    c_{\text{stbd}} = \frac{T_{\text{stbd},0}}{T_{\text{port},0}}
    \label{eq:c_stbd}
\end{equation}
\end{minipage}%
\hfill
\begin{minipage}{0.48\columnwidth}
\begin{equation}
    c_{\text{rev}} = \frac{K_{0,\text{rev}}}{K_{0,\text{fwd}}}
    \label{eq:c_rev}
\end{equation}
\end{minipage}
\end{subequations}
\vspace{5pt}

During yaw rotation $r$, the advance inflow velocity $u_{a,i}$ at thruster $i \in \{\text{port}, \text{stbd}\}$ depends on its lateral position $y_i$:
\begin{equation}
    u_{a,i} = u - y_i r
    \label{eq:u_adv_rot}
\end{equation}
where $y_{\text{port}}\!= -\frac{B}{2}$ and $y_{\text{stbd}}\!= +\frac{B}{2}$ with vessel beam width $B$.

Finally, the net force $T_{\text{force}, i}$ generated by thruster $i$ applies the advance inflow loss $u_{a,i}$ during forward operation and the constant efficiency scaling $c_{\text{rev}}$ during reverse operation:
\begin{equation}
    T_{\text{force}, i} = c_{\text{base}, i} T_{\text{act}, i} \cdot
    \begin{cases}
        \max(0, 1 - c_u u_{a, i}) & \text{if } T_{\text{act}, i} \geq 0 \\
        c_{\text{rev}} & \text{if } T_{\text{act}, i} < 0
    \end{cases}
    \label{eq:thrust_force}
\end{equation}

\vspace{-5pt}
\subsection{Real-World System Identification}
\label{sec:real_world_sysid}
The parameters ultimately fine-tuned in the simulation include hydrodynamic damping ($X_u, X_{uu}, Y_v, Y_{vv}, N_r, N_{rr}$), yaw inertia ($I_z$), and thruster coefficients $c_u$~\eqref{eq:c_u}, $c_{\text{rev}}$~\eqref{eq:c_rev}, and $\tau_{\text{lag}}$~\eqref{eq:c_lag}. For the real-world analytical baseline, the hydrodynamic damping and thruster coefficients, with the exception of $\tau_{\text{lag}}$ (nominally set to $0.10$\,s), are estimated using trajectories obtained from a physical BlueBoat in open-water conditions with minimal environmental disturbances. The remaining baseline parameters are established physically: mass $m$ is measured directly, while yaw inertia ($I_z$) and added mass terms ($X_{\dot{u}}$, $Y_{\dot{v}}$, $N_{\dot{r}}$) are obtained from a CAD model using a BEM solver.

\subsubsection{3-DOF Integration}
Assuming minimal wave-induced forces ($\boldsymbol{\tau}_{\text{wave}}=0$), geometric port-starboard symmetry, and negligible cross-coupling hydrodynamic forces at our low-speed operating regime, the 3-DOF dynamics~\eqref{eq:fossen_3dof} simplify to:
\begin{equation}
    M\dot{\boldsymbol{\nu}} + C(\boldsymbol{\nu})\boldsymbol{\nu} + D(\boldsymbol{\nu})\boldsymbol{\nu} = \boldsymbol{\tau}
\label{eq:fossen_sysid}
\end{equation}
By defining $m_{11} = m-X_{\dot{u}}$, $m_{22} = m-Y_{\dot{v}}$, and $m_{33} = I_z-N_{\dot{r}}$, the model matrices can be explicitly written as:
\begin{equation} 
\begin{split} 
M &= \text{diag}\big(m_{11},\; m_{22},\; m_{33}\big) \\
C(\boldsymbol{\nu}) &= \begin{bmatrix} 0 & 0 & -m_{22}v \\ 0 & 0 & m_{11}u \\ m_{22}v & -m_{11}u & 0 \end{bmatrix} \\ 
D(\boldsymbol{\nu}) &= -\text{diag}(X_u \!+\! X_{uu}|u|, Y_v \!+\! Y_{vv}|v|, N_r \!+\! N_{rr}|r|)
\label{eq:fossen_matrices1}
\end{split} 
\end{equation}
Expanding~\eqref{eq:fossen_sysid} yields the body-frame surge velocity $u$, sway velocity $v$, and yaw rate $r$ with $\boldsymbol{\tau} = \begin{bmatrix} \tau_X & \tau_Y & \tau_N \end{bmatrix}^\top$:
\begin{subequations}
\begin{align}
    (m-X_{\dot{u}})\dot{u} &= X_u u + X_{uu}u|u| + (m-Y_{\dot{v}})vr + \tau_X
    \label{eq:surge_vel} \\
    (m-Y_{\dot{v}})\dot{v} &= Y_v v + Y_{vv}v|v| - (m-X_{\dot{u}})ur + \tau_Y
    \label{eq:sway_vel} \\
    (I_z-N_{\dot{r}})\dot{r} &= N_r r + N_{rr}r|r| + (Y_{\dot{v}} - X_{\dot{u}})uv + \tau_N
    \label{eq:yaw_rate}
\end{align}
\end{subequations}

\subsubsection{Parameter Extraction}
Parameters are estimated via linear least squares using telemetry for a twin-screw ASV across a forward straight-line, backward straight-line, differential turn, and a circular single-thrust trajectory.

\textbf{Surge (}$X_u, X_{uu}, c_u, c_{\text{rev}}$\textbf{):} During straight-line coasting ($\tau_X = 0, v\approx0, r\approx0$), surge velocity~\eqref{eq:surge_vel} at time $t$ reduces to $(m-X_{\dot{u}})\dot{u}(t) = X_u u(t) + X_{uu}u(t)|u(t)|$, isolating linear and non-linear damping terms. During steady-state surge ($\dot{u}(t) = 0, v\approx0, r\approx0$), net surge thrust $\tau_X$ balances hydrodynamic drag $D_X (u) \!=\! -(X_u u \!+\! X_{uu}u|u|)$ derived from~\eqref{eq:surge_vel}. By substituting $\tau_X = T_{\text{cmd,total}}(1 - c_u u)$, where $T_{\text{cmd,total}}$ is the sum of commanded thrust, we solve for the advance loss $c_u$ and reverse efficiency $c_{\text{rev}}$ during steady-state forward and reverse operation, respectively.

\textbf{Yaw (}$N_r, N_{rr}$\textbf{):} During a counter-clockwise differential turn (forward starboard, reverse port), an active net yaw torque $\tau_N = \frac{B}{2} \left( T_{\text{force},\text{stbd}} - T_{\text{force},\text{port}} \right)$ across beam width $B$ is generated. Rearranging~\eqref{eq:yaw_rate} isolates the hydrodynamic yaw damping $D_N(r) = N_r r + N_{rr} r|r|$ against inertial, actuator, and Munk moment forces, yielding the linear and non-linear yaw coefficients, $D_N(r) = (I_z - N_{\dot{r}})\dot{r} - (Y_{\dot{v}} - X_{\dot{u}})uv - \tau_N$.

\textbf{Sway (}$Y_v, Y_{vv}$\textbf{):} A circular trajectory with aligned, single-thruster excitation ($\tau_Y = 0$) isolates sway dynamics. Rearranging~\eqref{eq:sway_vel} yields the linear and non-linear sway coefficients, $(m - Y_{\dot{v}})\dot{v} + (m - X_{\dot{u}})ur = Y_v v + Y_{vv}v|v|$.

\subsection{Simulation System Identification}
\label{sec:sim_sysid}

While the analytical SysID parameters derived in Section~\ref{sec:real_world_sysid} provide an accurate physical baseline, relying solely on them can increase the sim-to-real gap due to unmodeled hydrodynamics and physics engine approximations. To further ground the simulator's dynamics to the real-world ASV, we perform a secondary simulation-based SysID step using the covariance matrix adaptation evolution strategy (CMA-ES).

The optimization is framed as a sequential trajectory-matching problem utilizing 256 parallel simulation environments. To prevent over-parameterization and isolate dynamic effects, we replay the offline thruster commands from the four real-world trials sequentially, locking parameter subsets as they are identified. The surge forward trajectory isolates $X_u$, $X_{u|u|}$, and $c_u$; surge reverse identifies $c_{\text{rev}}$; differential turning isolates $I_z$, $N_r$, and $N_{r|r|}$; and the zigzag maneuver determines $Y_v$ and $Y_{v|v|}$. CMA-ES is warm-started using the physical baseline estimates, with the search space bounded to $\pm 30\%$ of these initial values. This constraint ensures the optimization yields physically plausible parameters and preserves the hydrodynamic linear-to-quadratic damping ratios, while providing sufficient flexibility to absorb simulator-specific deviations. 

To evaluate parameter fitness without falling into local minima caused by unbounded absolute errors during diverging trajectories, CMA-ES minimizes a continuous fitness function, $f$. Radial basis functions (RBFs) are utilized to map the state tracking errors into bounded similarity scores, $\phi \in (0, 1]$. The fitness function $f$ is defined as the negative sum of these similarity scores at each timestep $t$:
\begin{equation}
    f = - \sum_{t=1}^{T} \left( 1.5 \phi^{\text{pos}}_t + 0.5 \phi^{\psi}_t + 1.0 \phi^{\text{vel}}_t + 0.5 \phi^{r}_t \right)
\end{equation}
where the scalar coefficients denote the respective objective weightings. The individual RBF components evaluate the similarity between the simulated and recorded states based on a kernel width $\sigma$: $\phi^{\text{pos}}_t$ and $\phi^{\text{vel}}_t$ measure the deviation from the recorded $(x, y)$ path and $(u, v)$ linear velocities ($\sigma=1.0$), while $\phi^{\psi}_t$ and $\phi^{r}_t$ measure the deviation from the recorded yaw angle and yaw rate ($\sigma=0.75$).

Estimated parameter values for our physical BlueBoat obtained through real-world SysID alongside the final parameter values after simulation SysID are summarized in Table~\ref{tab:sysID_table}.

\subsection{Markov Decision Process Framework}

\subsubsection{Action Space}
\label{sec:action_space}
The action space at time step $t$ is defined as a continuous control vector $\mathbf{a}_t \in \mathbb{R}^N$, where $N \in \mathbb{N}_{\neq0}$ represents the number of propellers on the ASV (for the BlueBoat, $N=2$ with $i \in \{\text{port}, \text{stbd}\}$). Each element $a_{t,i}$ corresponds to the raw action allocated to the $i$-th propeller.

Raw actions $a_{t,i}$ are mapped to a normalized control space $x_{t,i} = \tanh\left(\frac{a_{t,i}}{5}\right)$, where $x_{t,i} \in (-1, 1)$. The normalized action is asymmetrically scaled to the real-world operational thrust range and filtered through a universal deadband threshold $f_{\text{dead}}$ to yield the target thrust command $T_{\text{cmd},i}[t]$:
\begin{equation}
    T_{\text{cmd},i}[t] = \begin{cases} 
        x_{t,i} \cdot |f_{\text{min},i}| & \text{if } x_{t,i} \cdot |f_{\text{min},i}| < -f_{\text{dead}} \\ 
        x_{t,i} \cdot f_{\text{max},i} & \text{if } x_{t,i} \cdot f_{\text{max},i} > f_{\text{dead}} \\
        0 & \text{otherwise}
    \end{cases}
\end{equation}
where $f_{\text{min},i}$ and $f_{\text{max},i}$ denote the reverse and forward force limits defined for the $i$-th propeller, respectively.

To account for the real-world thruster dynamics of the BlueBoat, this target command $T_{\text{cmd},i}[t]$ is evaluated through the physical thruster model defined in Section~\ref{sec:propeller_thrust}. Specifically, $T_{\text{cmd},i}[t]$ is subject to the motor spool-up lag~\eqref{eq:c_lag} to yield $T_{\text{act},i}$, which is then scaled by forward advance loss~\eqref{eq:c_u}, reverse operational efficiency~\eqref{eq:c_rev}, and starboard thrust asymmetry~\eqref{eq:c_stbd} to compute the net force $T_{\text{force}, i}$ applied to the vehicle~\eqref{eq:thrust_force}. The nominal action space parameters, including those obtained during SysID, are summarized in Table~\ref{tab:act_parameters}.

\begin{table}[!t]
\centering
\caption{Nominal Action Space Parameter Values}
\label{tab:act_parameters}
\begin{tabular}{lccc}
\hline
& \\[-2ex]
\textbf{Parameter} & \textbf{Port} & \textbf{Starboard} & \textbf{Units} \\ \hline
& \\[-2ex]
Reverse Thrust Limit ($f_{\text{min},i}$) & -10.0 & -10.0 & N \\
Forward Thrust Limit ($f_{\text{max},i}$) & 20.0  & 20.0  & N \\
Motor Force Deadband ($f_{\text{dead}}$) & \multicolumn{2}{c}{5.0} & N \\
Motor Spool-up Lag ($\tau_{\text{lag}}$) & \multicolumn{2}{c}{0.15} & s \\
Thrust advance loss ($c_u$) & \multicolumn{2}{c}{0.6016} & — \\
Reverse thrust coefficient ($c_{\text{rev}}$) & \multicolumn{2}{c}{0.4356} & — \\
Starboard Thrust Multiplier ($c_{\text{stbd}}$) & \multicolumn{2}{c}{1.0} & — \\ \hline
\end{tabular}
\end{table}

\subsubsection{Observation Space}
We employ an asymmetric actor-critic architecture, defining separate observation spaces for the policy (actor) and critic.

\paragraph{Policy Observation Space}
Both tasks, path following and station-keeping, share a core vehicle state vector $\mathbf{o}^{\text{base}, \pi}_t$:
\begin{equation}
    \mathbf{o}^{\text{base}, \pi}_t = \big[ u_t, v_t, r_t, x^{\text{acc}}_t, y^{\text{acc}}_t, \mathbf{a}^{\text{proc}}_{\text{hist}} \big]
\end{equation}
where $u_t$ and $v_t$ are the body-frame surge and sway velocities (m/s), $r_t$ is the yaw rate (rad/s), $x^{\text{acc}}_t$ and $y^{\text{acc}}_t$ are the body-frame surge and sway accelerations (m/s$^2$), and $\mathbf{a}^{\text{proc}}_{\text{hist}}$ represents a flattened history of five previously processed action vectors sampled with a stride of two.

The path following (PF) observation space is defined as:
\begin{equation}
    \mathbf{o}^{\text{PF}, \pi}_t = \big[ \mathbf{o}^{\text{base}, \pi}_t, \mathbf{p}^{\text{wp}}_t, e^{\text{cross}}_t, \sin(e^{\psi, \text{path}}_t), \cos(e^{\psi, \text{path}}_t) \big]
    \label{eq:obs_pf}
\end{equation}
where $\mathbf{p}^{\text{wp}}_t$ is a vector of the body-frame x and y errors to the closest waypoint and two subsequent waypoints, and $e^{\text{cross}}_t$ is the cross-track error. To eliminate $\pm\pi$ wrapping discontinuities, the path-relative heading error $e^{\psi, \text{path}}_t$ is embedded as continuous sine and cosine components.

The station-keeping (SK) observation space is defined as:
\begin{equation}
    \mathbf{o}^{\text{SK}, \pi}_t = \big[ \mathbf{o}^{\text{base}, \pi}_t, d_t, e^x_t, e^y_t, \sin(e^{\psi, \text{target}}_t), \cos(e^{\psi, \text{target}}_t) \big]
    \label{eq:obs_sk}
\end{equation}
where $d_t$ is the Euclidean (L2) distance to the target, and $e^x_t$ and $e^y_t$ are the decoupled body-frame surge and sway position errors. Similarly, the target-relative heading error $e^{\psi, \text{target}}_t$ is embedded as continuous sine and cosine components.

To simulate real-world sensor inaccuracies, the terms comprising $\mathbf{o}^{\text{PF}, \pi}_t$, $\mathbf{o}^{\text{SK}, \pi}_t$, and $\mathbf{o}^{\text{base}, \pi}_t$ are subject to uniform sensor noise (detailed in Section~\ref{sec:observation_noise}).

\paragraph{Critic Observation Space}
The critic is granted access to privileged environmental information along with noise-free states of the actor, $\mathbf{o}^{\text{task}, \pi}_t$, where $\text{task} \in \{ \text{PF}, \text{SK} \}$. The critic observation vector at time step $t$ for each task, denoted as $\mathbf{o}^{\text{task}, V}_t$, is defined as:
\begin{equation}
\begin{split}
    \mathbf{o}^{\text{task}, V}_t = \big[&
    \mathbf{o}^{\text{task}, \pi}_t,
    F^{h,x}_{t},
    F^{h,y}_{t},
    F^{w,x}_{t},
    F^{w,y}_{t},
    \sin(\gamma^b_t),
    \cos(\gamma^b_t) \big]
    \label{eq:obs_critic}
\end{split}
\end{equation}
where $F^{h,x}_{t}$ and $F^{h,y}_{t}$ represent the hydrodynamic forces acting on the vessel along the surge and sway axes of the ASV's body frame (N). $F^{w,x}_{t}$ and $F^{w,y}_{t}$ represent the true net aerodynamic wind forces acting on the vessel along the same axes (N). $\gamma^b_t$ represents the true wind direction relative to the ASV's current heading in the body frame (rad) embedded as sine and cosine components.

\begin{table}[!t]
\centering
\caption{Reward Function Weights and Hyperparameters}
\label{tab:rew_weights}
\begin{tabular}{lcl}
\hline
& \\[-2ex]
\textbf{Reward Component} & \textbf{$w_k$} & \textbf{Hyperparameters} \\ \hline
& \\[-2ex]
Cross-Track Error ($r^\text{cross}_t$) & $2.0$ & $\sigma_{\text{cross}} = 0.3$ \\
PF Heading ($r^{\text{PF,head}}_t$) & $1.0$ & $\sigma_{\text{PF,head}} = 0.3$ \\
Path Progress ($r^\text{prog}_t$) & $1.5$ & $\sigma_{\text{prog}} = 0.25, \ u_{\text{target}} = 0.5$ \\
PF Term Penalty ($r^{\text{PF,term}}_t$) & $-10.0$ & $d^{\text{PF}}_{\text{term}} = 2.0$ \\
Target Distance ($r^\text{dist}_t$) & $2.0$ & $\sigma_{\text{dist}} = 0.3$ \\
SK Heading ($r^{\text{SK,head}}_t$) & $1.0$ & $\sigma_{\text{SK,head}} = 0.3, \ \epsilon_{\text{SK,head}} = 2^{\circ}$ \\
SK Term Penalty ($r^{\text{SK,term}}_t$) & $-10.0$ & $d^{\text{SK}}_{\text{term}} = 1.0$ \\ \hline
\end{tabular}
\end{table}

\subsubsection{Reward Objective Functions}
The total reward $R_t$ at time step $t$ for each task is a combination of task-specific objectives, action-shaping, and termination conditions with weighting coefficients $w_k$.

To promote smooth control and minimize physical wear on the ASV's propellers during policy transfer, both tasks share shaping penalties applied to the thruster commands' first and second derivatives:
\begin{equation}
    r_t^{\text{act}} = -1.5\|\dot{\mathbf{a}}_t\|_2 -0.5\|\ddot{\mathbf{a}}_t\|_2
\end{equation}
Additionally, to incentivize high-precision tracking while maintaining a smooth gradient at larger errors, all continuous task-specific objectives use RBF Gaussian kernels.

The primary objective of the path following (PF) task is to maintain a target forward velocity while minimizing cross-track and heading error relative to a generated path:
\begin{equation}
    R^{\text{PF}}_t = w_1 r^\text{cross}_t + w_2 r^\text{PF,head}_t + w_3 r^\text{prog}_t + w_4 r^{\text{PF,term}}_t + r_t^{\text{act}}
\label{eq:rew_pf}
\end{equation}
The termination penalty $r^{\text{PF,term}}_t$ is applied if the ASV exceeds the cross-track limit ($|e^{\text{cross}}_t| > d^{\text{PF}}_{\text{term}}$). The continuous RBF objectives are defined as:
\begin{equation}
\begin{aligned}
    r^\text{cross}_t \!&=\! \exp\left( \!\!-\frac{(e^{\text{cross}}_t)^2}{2\sigma_{\text{cross}}^2} \!\right), \;\;
    r^\text{PF,head}_t \!=\! \exp\left( \!\!-\frac{(e^{\psi, \text{path}}_t)^2}{2\sigma_{\text{PF,head}}^2} \right) \\
    r^\text{prog}_t \!&=\! \exp\left( \!\!-\frac{(u^{\text{path}}_t - u_{\text{target}})^2}{2\sigma_{\text{prog}}^2} \right)
\end{aligned}
\end{equation}
where $e^{\text{cross}}_t$ and $e^{\psi, \text{path}}_t$ are the cross-track and path-relative heading errors introduced alongside~\eqref{eq:obs_pf}. $u^{\text{path}}_t$ is the ASV's velocity along the path tangent, and $u_{\text{target}}$ is the target velocity.

The primary objective of the station-keeping (SK) task is to minimize positional and heading deviations relative to the starting state despite external disturbances:
\begin{equation}
    R^{\text{SK}}_t = w_5 r^\text{dist}_t + w_6 r^\text{SK,head}_t + w_7 r^{\text{SK,term}}_t + r_t^{\text{act}}
\label{eq:rew_sk}
\end{equation}
The termination penalty $r^{\text{SK,term}}_t$ is applied if the ASV drifts beyond the allowed radius ($d_t > d^{\text{SK}}_{\text{term}}$). The continuous RBF objectives are defined as:
\begin{equation}
\begin{aligned}
    r^\text{dist}_t &= \exp\left( -\frac{d_t^2}{2\sigma_{\text{dist}}^2} \right) \\
    r^\text{SK,head}_t &= \exp\left( -\frac{\max(|e^{\psi, \text{target}}_t| - \epsilon_{\text{SK,head}}, 0)^2}{2\sigma_{\text{SK,head}}^2} \right)
\end{aligned}
\end{equation}
where $d_t$ and $e^{\psi, \text{target}}_t$ are the Euclidean (L2) distance to the target and target-relative heading error introduced alongside~\eqref{eq:obs_sk}. The heading term $r^\text{SK,head}_t$ incorporates a deadband threshold $\epsilon_{\text{SK,head}}$ to allow minor deviations without penalty.

The weighting coefficients $w_k$ and their associated hyperparameters for both tasks are summarized in Table~\ref{tab:rew_weights}.

\subsection{Domain Randomization}
To bridge the sim-to-real gap and train a policy robust against physical variances and environmental disturbances, we apply domain randomization during training. The domain randomization parameters used are summarized in Table~\ref{tab:dr_parameters}.

\subsubsection{Dynamics}
To simulate real-world drift at initialization, a uniformly sampled starting twist is applied to each ASV. Both tasks initialize with a surge velocity, sway velocity, and heading rate uniformly sampled from $[-0.2, 0.2]$~m/s, $[-0.1, 0.1]$~m/s, and $[\tfrac{-3\pi}{180}, \tfrac{3\pi}{180}]$~rad/s, respectively. For station-keeping, the initial pose is further randomized with a position offset within a $0.5$~m Manhattan distance and a heading error up to $\tfrac{30\pi}{180}$. Action space parameters modeling propeller thrust (defined in Section~\ref{sec:propeller_thrust}), with the exception of thrust limits $f_{\text{min},i}$ and $f_{\text{max},i}$, are sampled uniformly at the start of each episode. Hydrodynamic parameters calculated in Section~\ref{sec:real_world_sysid} are likewise sampled uniformly each episode.

\subsubsection{Wind Disturbance}
Simulated wind is applied to the ASV during training to build policy robustness to temporally correlated disturbances.
During each physics update, the ambient wind speed and direction evolve over time via an Ornstein-Uhlenbeck (OU) process. Across both path-following and station-keeping tasks, the initial wind direction is uniformly sampled across $[-\pi, \pi]$.
To facilitate learning, a curriculum schedule controls wind magnitude:
between steps 8,000 and 16,000, the initial wind speed range is linearly scaled from $[0, 0]$~m/s to $[0, 5]$~m/s for PF and $[0, 7]$~m/s for SK.


\subsubsection{Wave Disturbance}
Wave generation is introduced via a curriculum schedule after 64,000 steps. Both tasks share the same wave field configuration, defined by peak enhancement factor $\gamma = 3.3$, water depth $H = 50$~m, wind speed $4.0$~m/s, fetch $100$~km, and directional spreading factor $0.75$.

\begin{table}[!t]
\centering
\caption{Domain Randomization Parameters}
\label{tab:dr_parameters}
\begin{tabular}{lc}
\hline
& \\[-2ex]
\textbf{Parameter} & \textbf{Distribution} \\ \hline
& \\[-2ex]
Hydrodynamic multiplier & $\mathcal{U}(0.9, 1.1)$ \\
Advance loss ($c_u$) multiplier & $\mathcal{U}(0.9, 1.1)$ \\
Reverse efficiency ($c_u$) multiplier & $\mathcal{U}(0.9, 1.1)$ \\
Motor spool-up ($\tau_\text{lag}$) & $\mathcal{U}(0.10, 0.20)$ \\
Thrust asymmetry ($c_\text{stbd}$) & $\mathcal{U}(0.95, 1.05)$ \\
PF wind speed (m/s) & Curriculum: $\mathcal{U}(0, 0) \rightarrow \mathcal{U}(0, 5)$ \\
PF wind direction (rad) & $\mathcal{U}(-\pi, \pi)$ \\
SK wind speed (m/s) & $\mathcal{U}(0, 7)$ \\
SK wind direction (rad) & Curriculum: $\{0, \pi\} \rightarrow \mathcal{U}(-\pi, \pi)$ \\
Wind speed update (m/s) & OU ($\sigma=0.025, \theta=0.0025$) \\
Wind direction update (deg) & OU ($\sigma=0.75^{\circ}, \theta=0.0025$) \\ \hline
\end{tabular}
\end{table}

\subsubsection{Observation Noise}
\label{sec:observation_noise}
To emulate real-world sensor inaccuracies, additive Gaussian noise is applied to all components of the actor's observation vectors ($\mathbf{o}^{\text{PF}, \pi}_t$ and $\mathbf{o}^{\text{SK}, \pi}_t$), with the exception of the processed action history. The critic observations ($\mathbf{o}^{\text{task}, V}_t$) remain uncorrupted to provide the value function with the true privileged state.
Observation terms relating to position ($\mathbf{p}^{\text{wp}}_t, e^{\text{cross}}_t, d_t, e^x_t, e^y_t$) have a standard deviation of 0.025~m, while velocity ($u_t, v_t$) and acceleration ($x^{\text{acc}}_t$, $y^{\text{acc}}_t$) have standard deviations of 0.01~m/s and 0.02~m/s$^2$ respectively. Lastly, heading error ($e^{\psi, \text{path}}_t, e^{\psi, \text{target}}_t$) and heading rate ($r_t$) both have standard deviations of $1^\circ$.



\subsection{Training Network}
Policies are trained using Proximal Policy Optimization (PPO)~\cite{schulman2017ppo} implemented via the skrl~\cite{serrano2023skrl} reinforcement learning library. The training process is parallelized across multiple independent environments using a physics and hydrodynamics simulation time step of 1/120~s and a policy query frequency of 60~Hz. The policy (actor) and value (critic) networks are both Multilayer Perceptrons (MLPs) with Exponential Linear Unit (ELU) activations. Following the asymmetric actor-critic architecture, separate weights are maintained for each network. Actor observations, critic states, and computed value targets are standardized by removing the mean and scaling to unit variance.
Both tasks were trained in 256 parallel environments over 1000 iterations using shared hyperparameters: 5 learning epochs, 16 mini-batches, a $128 \times 64$ hidden layer architecture, learning rate of $3.0 \times 10^{-4}$, GAE $\lambda = 0.95$, KL threshold and entropy coefficient of 0.01, surrogate and value clipping of 0.20, and gradient norm clipping of 0.50. Parameters that differ between tasks are listed in Table~\ref{tab:ppo_hyperparameters}.

\begin{table}[!t]
\centering
\caption{Task-Specific Training Hyperparameters}
\label{tab:ppo_hyperparameters}
\begin{tabular}{lcc}
\hline
\textbf{Parameter} & \textbf{Path Following (PF)} & \textbf{Station Keeping (SK)} \\ \hline
Episode Duration & 20~s & 30~s \\
Rollout Steps & 128 & 256 \\
Discount Factor ($\gamma$) & 0.99 & 0.9975 \\
\hline
\end{tabular}
\end{table}


\begin{figure*}
    \centering
    \includegraphics[width=1.0\linewidth]{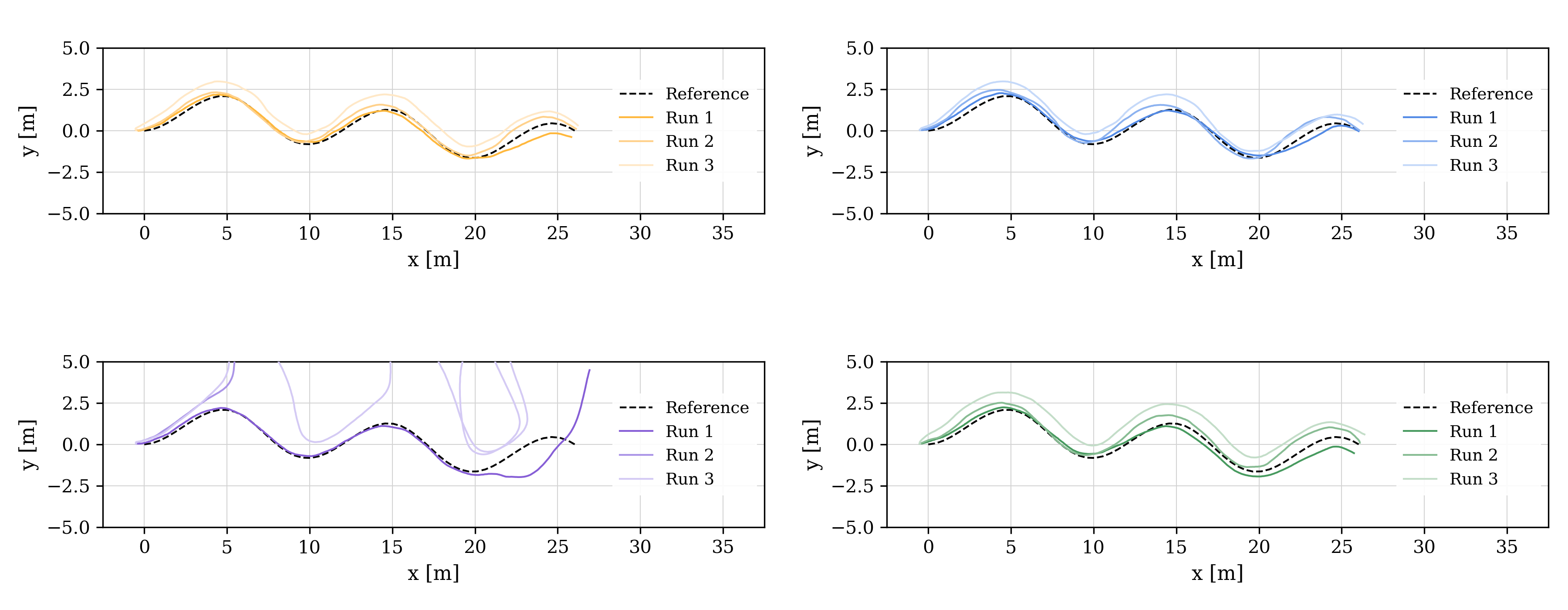}
    \caption{Real-world path following trajectory evaluations for each configuration: \textbf{Real-World} (top-left), \textbf{Proposed} (top-right), \textbf{Zigzag} (bottom-left), and \textbf{CMA-ES DR} (bottom-right). Proposed achieves the lowest average cross-track RMSE and single-trajectory grounding suffers from divergent dynamics.}
    \label{fig:pf_eval}
\end{figure*}

\section{Experiments}

To evaluate the impact of our simulation grounding and isolate the CMA-ES curriculum's contribution, we train and evaluate RL policies across four distinct configurations:
\begin{itemize}
    \item Analytical SysID \textbf{(Real-World)}: Uses the parameter estimates derived in Section~\ref{sec:real_world_sysid} without CMA-ES.
    \item Simulation Grounding \textbf{(Proposed)}: Uses grounded simulator parameters derived via the sequential trajectory-matching CMA-ES curriculum outlined in Section~\ref{sec:sim_sysid}.
    \item Single-Trajectory Grounding \textbf{(Zigzag)}: An ablation configuration attempting to optimize all nine parameters simultaneously using only the zigzag maneuver while using the same real-world priors and $\pm 30\%$ bounds.
    \item CMA-ES Domain Randomization \textbf{(CMA-ES DR)}: While using the same dynamics values as the proposed method, adaptive DR scaling proportional to the optimized parameter's proximity to its CMA-ES search bounds is applied. Using the magnitude of the raw CMA-ES decision variable $|x| \in [0, 3]$, the DR multiplier is sampled from a uniform distribution $\mathcal{U}(1 - 0.10|x|, 1 + 0.10|x|)$. This dynamically scales the parameter variance up to $\pm 30\%$, injecting broader randomization into parameters that deviated further from the real-world priors.
\end{itemize}


\begin{table}[h]
\centering
\caption{SysID Parameters Across Experimental Configurations}
\label{tab:sysID_table}
\begin{tabular}{lcccc}
\hline
\\[-2ex]
\textbf{Parameter} & \textbf{Real-World} & \textbf{Proposed} & \textbf{Zigzag} & \textbf{Units} \\ \hline
\\[-2ex]
$I_z$ & 4.86 & 6.31 & 4.96 & kg$\cdot$m$^2$ \\
$X_u$ & -0.46 & -0.59 & -0.59 & kg/s \\
$X_{uu}$ & -6.83 & -6.98 & -8.87 & kg/m \\
$Y_v$ & -54.89 & -71.26 & -71.28 & kg/s \\
$Y_{vv}$ & -114.30 & -148.37 & -148.42 & kg/m \\
$N_r$ & -6.38 & -4.48 & -8.29 & kg$\cdot$m$^2$/s \\
$N_{rr}$ & -18.76 & -13.16 & -24.36 & kg$\cdot$m$^2$ \\
$c_u$ & 0.602 & 0.648 & 0.781 & — \\
$c_{\text{rev}}$ & 0.436 & 0.539 & 0.566 & — \\\hline
\end{tabular}
\end{table}


At inference time, the policy is evaluated via a ROS 2 node streaming thrust commands to the BlueBoat.
The neural network architecture and actor observation terms remain identical to training, populated using live sensor telemetry. Field trials were conducted in a river setting under calm weather conditions. To evaluate consistency under hydrodynamic disturbances, each policy underwent three runs for path following and four runs for station-keeping, with each run initialized from a different starting pose to expose the system to varying localized currents. To ensure a fair comparison, the same sets of initial poses were reused across all evaluated policies.

\begin{figure}
    \centering
    \includegraphics[width=1.0\linewidth]{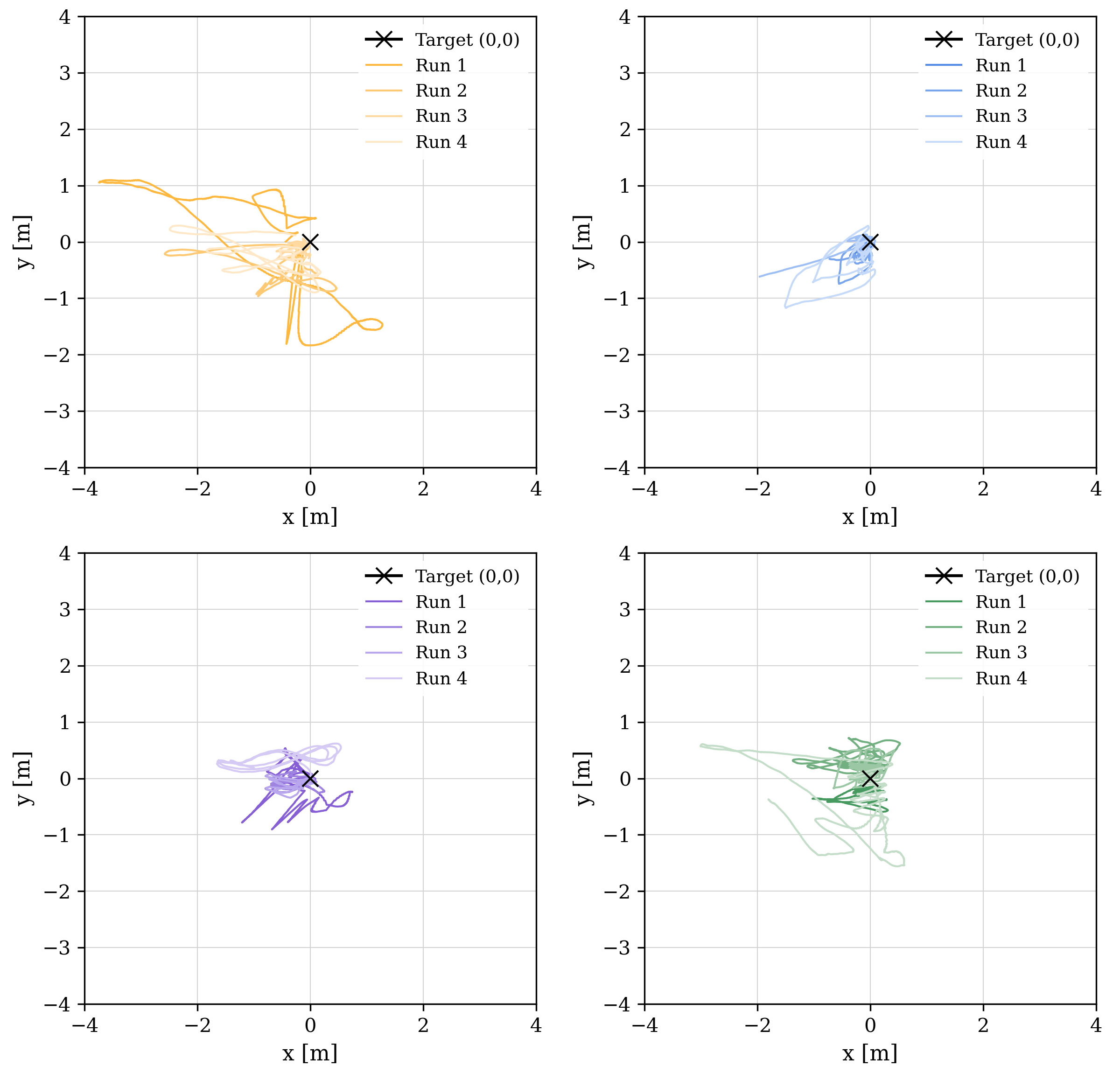}
    \caption{Real-world station-keeping evaluations lasting 45~s for each configuration: \textbf{Real-World} (top-left), \textbf{Proposed} (top-right), \textbf{Zigzag} (bottom-left), and \textbf{CMA-ES DR} (bottom-right). Grounded policies all similarly outperform Real-World, with Zigzag notably outperforming CMA-ES DR.}
    \label{fig:sk_eval}
\end{figure}

\section{Results and Discussion}

This section evaluates the real-world performance of trained policies, with the average root mean squared error (RMSE) and mean absolute error (MAE) summarized in Table~\ref{tab:eval}.

\textbf{Proposed Configuration Accuracy}: The Proposed configuration achieved the lowest tracking error across all tasks. For path following (PF), it achieved a $0.417$~m cross-track RMSE, slightly outperforming the direct Real-World configuration ($0.423$~m RMSE). Notably, during station-keeping (SK), the Proposed policy reduced position RMSE by $63.9\%$ compared to the Real-World configuration (1.074~m to 0.388~m RMSE).

\begin{table}[!t]
\centering
\caption{Real-World Evaluation RMSE and MAE}
\label{tab:eval}
\begin{tabular}{lcccc}
\hline
\\[-2ex]
\textbf{Evaluation Task} & \!\!\!\textbf{Real-World}\!\!\! & \!\!\textbf{Proposed}\!\! & \!\!\textbf{Zigzag}\!\! & \!\!\!\textbf{CMA-ES DR}\!\!\! \\ \hline
\\[-2ex]
PF Cross-Track RMSE\!\!\! & 0.423~m & 0.417~m & \!\!2.909~m\!\! & 0.549~m \\
SK Position RMSE & 1.074~m & 0.388~m & \!\!0.569~m\!\! & 0.643~m \\
SK Heading MAE & 17.30$^\circ$ & 10.37$^\circ$ & 12.70$^\circ$ & 10.68$^\circ$ \\\hline
\end{tabular}
\end{table}

\textbf{Single-Trajectory Grounding}:
Despite achieving comparable training rewards to other configurations, the Zigzag ablation proved unstable in real-world testing after missing a turn. While it achieved reasonable position RMSE during station-keeping, it yielded the worst heading MAE among grounded methods. This demonstrates that incomplete dynamic grounding causes brittle performance during maneuvers reliant on those missing dynamics, such as heading changes.

\textbf{Effect of CMA-ES Adaptive DR}: Introducing CMA-ES DR degraded tracking performance in both tasks compared to the Proposed configuration. CMA-ES DR scales randomization based on a parameter's deviation from its real-world prior, resulting in heavy variance for significantly shifted sway and yaw damping parameters ($\pm 26.95\%$ to $\pm 30.0\%$) and negligible variance for $c_u$ ($\pm 2.6\%$).
Consequently, the disproportionate application of variance failed to improve policy robustness. Compared to our Proposed configuration, CMA-ES DR demonstrated increased lateral drift in real-world currents for both the path-following and station keeping tasks.






\section{Conclusion}
This paper presents an ASV simulator with hydrodynamics, waves, and thruster modeling, along with a real-to-sim-to-real pipeline for zero-shot deployment of RL policies. Using a set of four open-water trajectories, we ground simulator dynamics and evaluate our pipeline on a real-world BlueBoat in path following and station-keeping tasks.
Future work will focus on the importance of individual hydrodynamic terms and their domain randomization for sim-to-real transfer.

\bibliographystyle{IEEEtran}
\bibliography{bibliography}

\end{document}